\documentclass[review]{elsarticle}

\usepackage{lineno,hyperref}

\usepackage{lineno,hyperref}

\usepackage{times}
\usepackage{epsfig}
\usepackage{graphicx}
\usepackage{amsmath}
\usepackage{amssymb}

\usepackage{url}
\usepackage{subfigure}

\usepackage{array}
\usepackage{multirow}
\usepackage{booktabs}
\usepackage{threeparttable}
\newcommand{\PreserveBackslash}[1]{\let\temp=\\#1\let\\=\temp}
\newcolumntype{C}[1]{>{\PreserveBackslash\centering}p{#1}}
\newcolumntype{R}[1]{>{\PreserveBackslash\raggedleft}p{#1}}
\newcolumntype{L}[1]{>{\PreserveBackslash\raggedright}p{#1}}

\modulolinenumbers[5]

\begin{document}

\begin{frontmatter}

\title{3D Face Mask Presentation Attack Detection Based on \\ Intrinsic Image Analysis}

\author[1]{Lei Li}
\author[1]{Zhaoqiang Xia}
\author[1]{Xiaoyue Jiang}
\author[1]{Yupeng Ma}
\author[2,1]{Fabio Roli}
\author[1]{Xiaoyi Feng}

\address[1]{School of Electronics and Information, Northwestern Polytechnical University, Xi'an, Shaanxi, China}
\address[2]{Department of Electrical and Electronic Engineering, University of Cagliari, Cagliari, Sardinia, Italy}

\begin{abstract}
  Face presentation attacks have become a major threat to face recognition systems and many countermeasures have been proposed in the past decade. However, most of them are devoted to 2D face presentation attacks, rather than 3D face masks. Unlike the real face, the 3D face mask is usually made of resin materials and has a smooth surface, resulting in reflectance differences. So, we propose a novel detection method for 3D face mask presentation attack by modeling reflectance differences based on intrinsic image analysis. In the proposed method, the face image is first processed with intrinsic image decomposition to compute its reflectance image. Then, the intensity distribution histograms are extracted from three orthogonal planes to represent the intensity differences of reflectance images between the real face and 3D face mask. After that, the 1D convolutional network is further used to capture the information for describing different materials or surfaces react differently to changes in illumination. Extensive experiments on the 3DMAD database demonstrate the effectiveness of our proposed method in distinguishing a face mask from the real one and show that the detection performance outperforms other state-of-the-art methods.
\end{abstract}


\end{frontmatter}


\section{Introduction}
Face presentation attack detection (PAD) has become an important issue in face recognition, since the biometric technologies have been applied in various verification systems and mobile devices. For instance, Apple's face ID for iPhone X has been used to unlock cell phones and complete mobile payments. However, D. Oberhaus \cite{Daniel2017iPhone} reported that some researchers at the Vietnamese cybersecurity firm Bkav had made a mask that could fool the Face ID system. Considering such urgent security situation, an effective and reliable 3D face mask PAD method must be developed.

\begin{figure*}[!t]
 \centering
 \includegraphics[width=0.9\textwidth,angle=0]{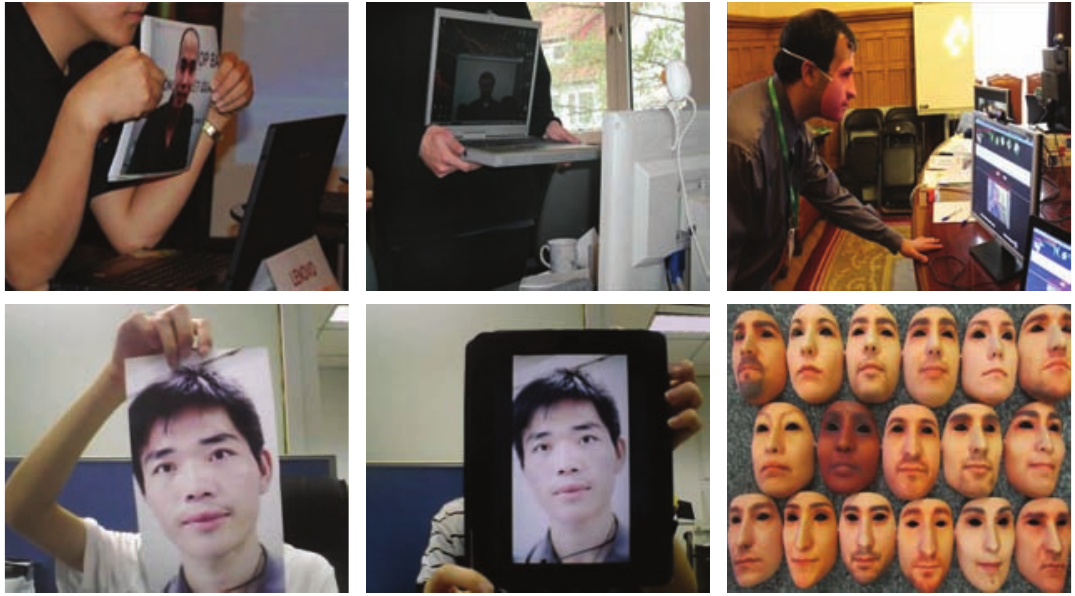}
 \caption{Samples of face presentation attacks. From left to right: printed face photos, displayed face images or replayed videos, and 3D masks.}\label{fig:differentAttacks}
\end{figure*}

Based on different presented fake faces, four kinds of presentation attacks can be considered: printed photo, displayed image, replayed video and 3D mask. The printed photo, displayed image and replayed video attacks usually print or display a face image in 2D medium (e.g., paper and LCD screen). But for 3D face mask scenario, the attacker wears a 3D mask to deceive face recognition system. Compared to 2D face presentation attacks, 3D face masks contain the structural and depth information similar to real faces. Fig. \ref{fig:differentAttacks} shows an example of different face presentation attacks.

In the last decade, many face PAD approaches have been proposed \cite{Chingovska2012On,Erdogmus2013Spoofing,Maatta2011Face,Boulkenafet2015Face,Pan2007Eyeblink,Li2009An,Zhang2012A,Tan2010Face,Li2017Face,Pavlidis2000The,Zhang2011Face,Li2018Face2}. However, the main focus of them has been on tackling the problem of 2D face presentation attacks, while 3D attacks have received much less attention. It is because detecting 3D face masks and distinguish them from real faces more challenging for common cameras. For instance, Liu \textit{et al.} \cite{Liu2018Learning} estimated face depth by a CNN-RNN model and used the depth information to detect presented fake faces. This method can solve the 2D face PAD successfully. Whereas, this detection method cannot work well in a 3D mask attack as the mask also has depth information. In another work, Li \textit{et al.} \cite{Li2017Generalized} tackled the 3D mask attacks by detecting the pulse from videos, which may be sensitive to camera settings and light conditions. Besides, it becomes easier to obtain 3D masks by attackers with the development of 3D printing. Thus, it is necessary to develop new methods for detecting 3D face mask.

\begin{figure*}[!t]
 \centering
 \includegraphics[width=1\textwidth,angle=0]{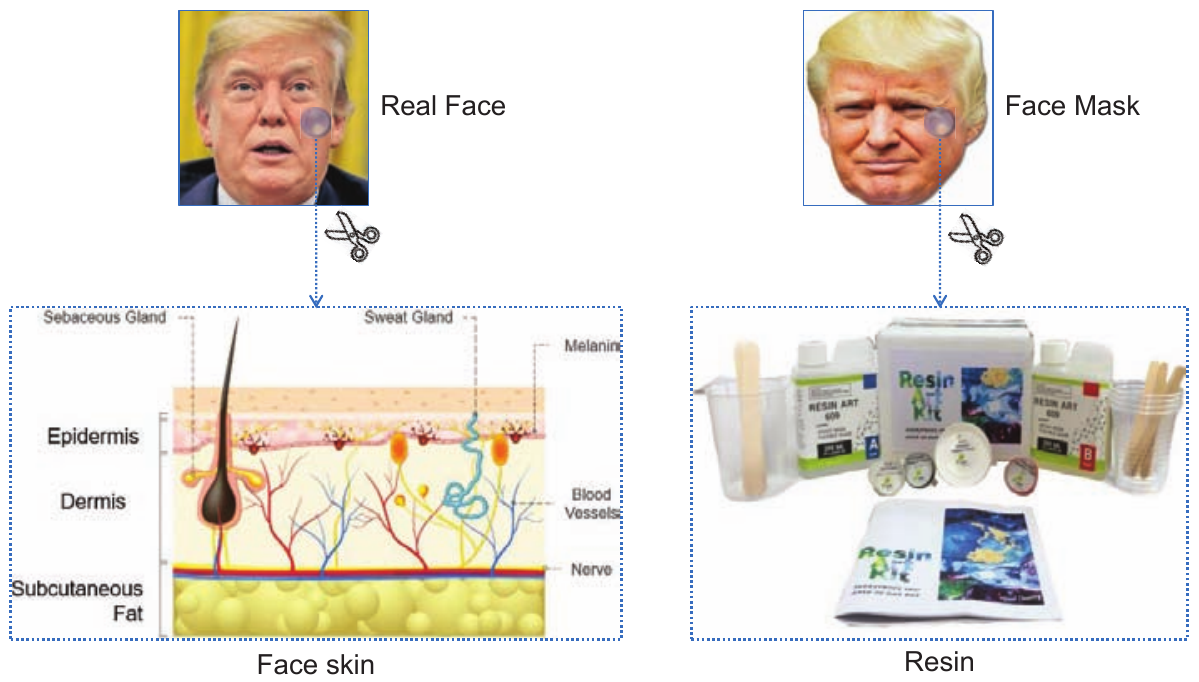}
 \caption{The difference between real face (left) and face mask (right) in skin material. For the real face skin, it consists of three layers: epidermis, dermis and subcutaneous fat. But for the face mask, it is often made of resin materials.}
 \label{fig:Face_skin_resin}
\end{figure*}

In this work, we propose to analyze the intrinsic image for 3D face mask PAD. Intrinsic image decomposition can divide a single image into a reflectance image and a shading image \cite{Ma2017Intrinsic}. For the reflectance image, it gives the albedo at each point, which describes the reflectance characteristics and is only decided by the material itself. In contrast, the shading image gives the total light flux incident at each point. Comparing the real face with face mask, we can find that their constituted materials are fairly different. Furthermore, due to the influence of hair follicles, the surface of 3D face mask is smoother than the surface of real face. More specifically, the face skin consists of three layers: epidermis, dermis and subcutaneous fat \footnote{https://www.webmd.com/skin-problems-and-treatments/ss/slideshow-skin-infections} as illustrated in Fig. \ref{fig:Face_skin_resin} (left). But for the 3D face mask, it is often made of resin materials as shown in Fig. \ref{fig:Face_skin_resin} (right). These material or surface differences can be revealed in reflectance images and further used for analyzing visual difference.

Our proposed method can be divided into three procedures: (i) the captured face image is first decomposed into a reflectance image and a shading image based on correlation-based intrinsic image extraction (CIIE) \cite{Jiang2010Correlation}; (ii) the intensity distribution histograms of reflectance images are calculated from three orthogonal planes and the 1D convolutional neural network (CNN) is introduced to extract the differences in reflectance images; (iii) the extracted features are fed into a Support Vector Machine (SVM) classifier to detect 3D face mask attack. The main contributions of this work are summarized:

$\diamond$ Unlike most previous works on 3D face mask PAD are based on analyzing visible cues (e.g. texture) directly, we propose a novel and appealing countermeasure from the view of intrinsic image decomposition.

$\diamond$ We design an intensity distribution feature for describing the difference of reflectance images between the real face and 3D face mask and introduce 1D CNN to represent different materials or surfaces react differently to the changes in variational illumination.

$\diamond$ We evaluate the proposed method on 3DMAD dataset and achieve the state-of-the-art performance.

\section{Related Work}

\subsection{Methods for 2D Face PAD}

\noindent\textbf{Texture-analysis based methods.} Due to the limitations of printers and display devices, the printed or replayed fake faces generally present blurred textures compared to the real ones. Thus, using texture information has been a natural approach to tackling 2D face presentation attack. For instance, many works employed hand-crafted features, such as local binary pattern (LBP) \cite{Maatta2011Face}, Haralick features \cite{Agarwal2016Face} and scale-invariant feature transform (SIFT) \cite{Patel2016Secure}, and then used traditional classifiers, such as SVM and LDA. To capture the color differences in chrominance, the texture features extracted from different color spaces (e.g. RGB, HSV and YCbCr) are also been analyzed \cite{Boulkenafet2016FaceTIFS}. Apart from that, some feature descriptors or indexes are computed to describe the clarity of facial texture \cite{Tan2010Face,Zhang2012A,Wen2015Face}. More recently, many attempts of using CNN-based features in face PAD \cite{Yang2014Learn,Li2017An,Li2018Face}. While these methods are effective to typical 2D paper or replayed attacks, they become vulnerable when attackers wear a lifelike face mask.

\noindent\textbf{Motion-analysis based methods.}
Apart from analyzing the texture differences, motion is another important clue for printed photo and displayed image attacks. The most commonly used solutions are based on detecting aliveness signals, such as eye blinking \cite{Pan2007Eyeblink,Sun2007Blinking}, head motion \cite{Anjos2011Counter}, facial muscle twitching \cite{Zhang2010Local} and pulse \cite{Liu2018Learning}. Besides, there are methods relying on the planar object movements \cite{Li2009An} and the optical flow fields \cite{Bao2009A}. Moreover, in order to capture the temporal texture variations from a video sequence as \cite{Pereira2014Face}, the deep networks with timing processing such as long short memory network (LSTM) and 3D CNN are used to detect face presentation attack \cite{Xu2016Learning,Li2018Learning}. These algorithms make it easy to detect the printed or displayed image attack. But when facing with 3D face mask attack with eye/mouth portion being cut, their performance will be greatly discounted.

\noindent\textbf{Hardware based methods.}
In recent years, various hardwares have been applied for the task of face PAD. For instance, the Kinect camera was used to capture depth information and detect fake faces \cite{Erdogmus2013Spoofing}. The reason is that there is no any depth information in 2D presentation attacks. Also, the near-infrared (NIR) sensors and photodiodes are explored to enhance the optical difference between the real and fake faces \cite{Pavlidis2000The,Zhang2011Face}. More recently, light-field cameras allow exploiting disparity and depth information from a single capture. Therefore, some works introduced these kinds of cameras into face PAD \cite{Kim2014Face,Ji2016LFHOG,Sepas2017Light}. Even though those hardware-based methods can achieve good performances for 2D face PAD, some of them might present operation restrictions in certain conditions. For instance, the sunlight can cause severe perturbations for NIR and depth sensors; wearable 3D masks are obviously challenging for those methods relying on depth data.

\subsection{Methods for 3D Face Mask PAD}
3D face mask presentation attack is a new threat to face biometric systems and few countermeasures have been proposed compared to conventional 2D attacks. To address this issue, Erdogmus \textit{et al.} \cite{Erdogmus2014Spoofing} released the first 3D mask attack dataset (3DMAD). Different from printed or replayed fake faces, the attackers in 3DMAD wear 3D face masks of targeted persons. Inspired by \cite{Maatta2011Face}, Erdogmus \textit{et al.} utilized multi-scale LBP based texture representation for 3D face mask PAD. In addition, Liu \textit{et al.} \cite{Liu20163D} presented an rPPG-based method to cope with 3D mask attacks. In another work, Li \textit{et al.} \cite{Li2017Generalized} tackled the 3D mask attacks based on pulse detection. These two methods mainly analyze the different heartbeat signals between real faces and 3D masks. However, they may be sensitive to camera settings and light conditions. More recently, Shao \textit{et al.} \cite{Shao2017Deep} extracted optical flow features from a pre-trained VGG-net \cite{Simonyan2014Very} to estimate subtle motion.

\section{Proposed Method}
\subsection{Rationale and Motivations}
\begin{figure*}[!ht]
\centering
\subfigure[Real face]{\includegraphics[width=0.45\textwidth]{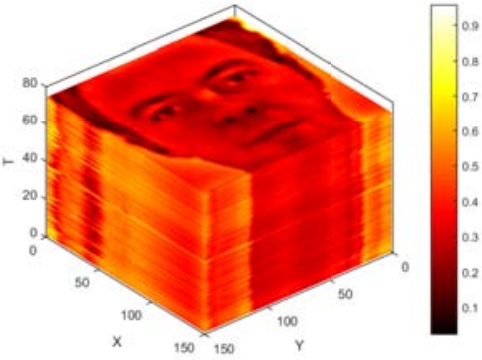}
\hfil
\label{fig:IntrinsicRealHot}}
\subfigure[3D face mask]{\includegraphics[width=0.45\textwidth]{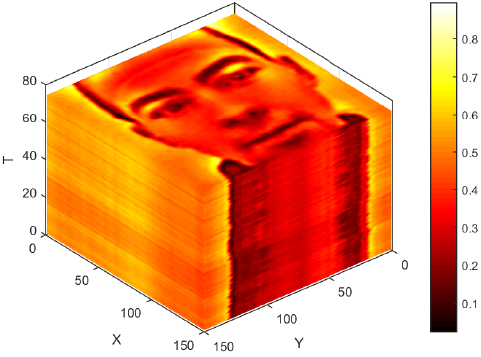}
\hfil
\label{fig:IntrinsicFakeHot}}
\caption{The hot figures of the reflectance image sequences of real face (left) and 3D face mask (right).}
\label{fig:IntrinsicDifferent}
\end{figure*}

As aforementioned, the constituted materials of real face and 3D face mask are different and these differences can be revealed in reflectance images, which are shown in Fig. \ref{fig:IntrinsicRealHot} and \ref{fig:IntrinsicFakeHot}. From the example, it can be found that the reflectance images of real face have low intensity and uniform distribution while the reflectance images of 3D face mask have high intensity and non-uniform distribution. Based on this observation, we propose a new method to discriminate the reflectance difference of real faces and 3D face masks.

In reflectance images of real and spoofing faces, reflectance images often present obvious intensity difference due to different materials and surface structures. To describe the intensity differences, we extract intensity histograms from the view of albedo intensity. Furthermore, the skin surface of the real face is rougher than the 3D face mask, which is caused by hair follicles, etc. When lighting changes or the face shake, the reflectance images of real face and 3D face mask also show different responses. For instance, the reflectance image of a smooth surface is more sensitive to lighting changes. So in addition to extracting the intensity histogram, we further use a 1D CNN to extract the difference in illumination sensitivity of the reflectance image.

\begin{figure*}[!ht]
 \centering
 \includegraphics[width=1\textwidth,angle=0]{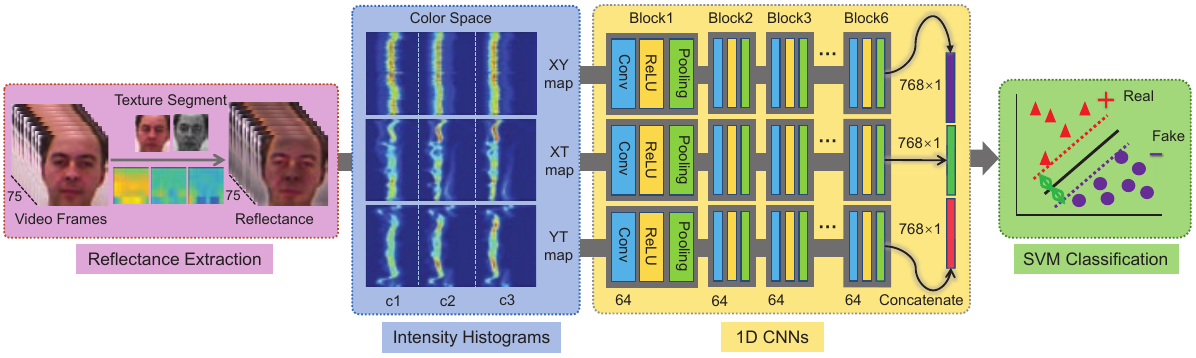}
 \caption{The flowchart of our proposed 3D face mask PAD method. It consists of four modules: Reflectance Extraction, Intensity Histograms, 1D CNNs and SVM classification. In the second module, XY, XT and YT maps are calculated from three orthogonal planes of different color channels (denoted as c1, c2 and c3). In 1D CNNs module, the channel number of convolutional feature maps is 64 and the dimension of the last layer is $768 \times 1$.}
 \label{fig:FlowChart}
\end{figure*}

Therefore, as shown in Fig. \ref{fig:FlowChart}, our proposed approach is divided into four modules, i.e., Reflectance Extraction, Intensity Histograms, 1D CNNs and SVM classification. First, the frames of captured face video are decomposed into reflectance images and shading images. Then, the intensity histograms of reflectance images are calculated from three orthogonal planes (TOP) and combined into three histogram feature maps, which can describe the ratio and distribution of different intensities. After that, these TOP histogram feature maps are fed into three 1D CNNs to extract the difference in illumination sensitivity, respectively. Finally, an SVM classifier is employed to distinguish whether the concatenated deep features are oriented from a valid access or a 3D face mask.

\subsection{Intrinsic Image Decomposition}
Based on the Lambertian theory \cite{Ma2017Intrinsic}, the intensity value $I(x,y)$ of every pixel in an image is the product of the shading $S(x,y)$ and the reflectance $R(x,y)$ at that point: $I(x,y)=S(x,y)\times R(x,y)$. The shading $S(x,y)$ represents illumination changes and is decided by external light intensity. $R(x,y)$ gives the albedo at each point. In addition, albedo completely describes the reflectance characteristics for Lambertian (perfectly diffusing) surfaces. Considering different materials of real face and 3D face mask, we will analyze the reflectance images of them. In the context, we utilize the CIIE algorithm \cite{Jiang2010Correlation} to decompose the original face image.

More specifically, the steerable filters are first used to decompose the face image into its constituent orientation/frequency bands. Then, the filter bank is applied to estimate the luminance components ($LM_{ij}$) and the variations in local amplitude ($AM$), texture ($TM$), and hue ($HM$), respectively. Based on this assumption that luminance variations are either due to reflectance or illumination but not both, the estimated shading and reflectance can be reconstructed as follows:
\begin{eqnarray}\label{Eq:Intrinsic_shadding}
  S=F_{S}\otimes \left\{
             \begin{array}{lr}
             C_{ij}^{rec\_ shd}\times LM_{ij} \quad & if C_{ij}^{rec\_ shd} > 0  \\
             0  \quad & if C_{ij}^{rec\_ shd} \leq 0
             \end{array}
\right.
\end{eqnarray}
\begin{eqnarray}\label{Eq:Intrinsic_reflectance}
  R=F_{S}\otimes \left\{
             \begin{array}{lr}
             C_{ij}^{rec\_ ref}\times LM_{ij} \quad & if C_{ij}^{rec\_ ref} > 0  \\
             0  \quad & if C_{ij}^{rec\_ ref} \leq 0
             \end{array}
\right.
\end{eqnarray}
where $ij$ is the index of $LM$ and $\otimes$ is the reconstruction of steerable filters $F_{S}$ with weighted $LM_{ij}$. $C_{ij}^{rec\_ shd}$ and $C_{ij}^{rec\_ ref}$ are correlation coefficients for reconstructed shading and reflectance, respectively.

During image reconstruction, some $LM$ components will be set to zero and resulting images may lose their DC values. Meanwhile, some $AM$ components will represent small variations in reflectance not shading. Thus two cost functions are used to optimize $S$ and $R$, as illustrated in Eq. \ref{Eq:Optimize_Shadding} and Eq. \ref{Eq:Optimize_Reflectance}.
\begin{eqnarray}\label{Eq:Optimize_Shadding}
  E_{S}(V_{DC})=\arg min_{T_{i}}\sum _{i}F_{txt}(\frac{I}{S+V_{DC}}, T_{i})
\end{eqnarray}
\begin{eqnarray}\label{Eq:Optimize_Reflectance}
  E_{R}=\arg min_{k}\sum _{i}F_{txt}(\hat{LM}_{ij}, T_{k}) + \lambda E_{cst}(F_{adj})
\end{eqnarray}
where $V_{DC}$ is the DC value to be optimized. The function $F_{txt}(\hat{I}, T_{i})$ represents texture consistency evaluated for image $\hat{I}$ against the texture segmentation result. $ \hat{LM}_{ij}$ is modelled as a normal distribution and $E_{cst}(F_{adj})$ constrains the maximum rescaling produced by the adjustment function. Fig. \ref{fig:IntrinsicDifferent} shows the reflectance image sequences of a real face and a 3D face mask.


\begin{table*}[!ht]\scriptsize
\centering
\caption{The configuration parameters in 1D CNN. \textit{aPool} is the operation of average pooling and \textit{Trans} is the operation of matrix transpose.}\label{tab:Detail-CNN}				
\begin{tabular}{|C{0.9cm}|C{0.5cm}|C{0.5cm}|C{0.5cm}|C{0.5cm}|C{0.5cm}|C{0.5cm}|C{0.5cm}|C{0.5cm}|C{0.5cm}|C{0.6cm}|C{0.5cm}|}
\hline
layer     &$1$      &$2$     &$3$     &$4$     &$5$   &$6$     &$7$      &$8$     &$9$    &$10$     &$11$\\
type      &Conv     &ReLU    &aPool   &Conv    &ReLU  &aPool   &Conv     &ReLU    &aPool  &Conv     &ReLU  \\
\hline\hline
filt size &$[3,1]$  &$-$     &$-$     &$[3,1]$ &$-$   &$-$     &$[3,1]$  &$-$     &$-$    &$[3,1]$  &$-$\\
filt dim  &$1$      &$-$     &$-$     &$64$    &$-$   &$-$     &$64$     &$-$     &$-$    &$64$     &$-$\\
num filts &$64$     &$-$     &$-$     &$64$    &$-$   &$-$     &$64$     &$-$     &$-$    &$64$     &$-$\\
stride    &$1$      &$1$     &$[2,1]$ &$1$     &$1$   &$[2,1]$ &$1$      &$1$     &$[2,1]$&$1$      &$1$\\
pad       &$[1,0]$  &$0$     &$0$     &$[1,0]$ &$0$   &$0$     &$[1,0]$  &$0$     &$0$    &$[1,0]$  &$0$\\
\hline
\hline
layer     &$12$     &$13$    &$14$    &$15$    &$16$    &$17$    &$18$     &$19$     &$20$    &$21$    &$22$\\
type      &aPool    &Conv    &ReLU    &aPool   &Conv    &ReLU    &Conv     &Trans    &FC      &SoftMax &$-$\\
\hline\hline
filt size &$-$      &$[3,1]$ &$-$     &$-$     &$[2,1]$ &$-$     &$[1,1]$  &$-$      &$[1,1]$ &$-$     &$-$\\
filt dim  &$-$      &$64$    &$-$     &$-$     &$64$    &$-$     &$64$     &$-$      &$64$    &$-$     &$-$\\
num filts &$-$      &$64$    &$-$     &$-$     &$64$    &$-$     &$64$     &$-$      &$2$     &$-$     &$-$\\
stride    &$[2,1]$  &$1$     &$1$     &$[2,1]$ &$1$     &$1$     &$1$      &$1$      &$1$     &$1$     &$-$\\
pad       &$0$      &$[1,0]$ &$0$     &$0$     &$0$     &$0$     &$0$      &$0$      &$0$     &$0$     &$-$\\
\hline
\end{tabular}
\end{table*}

\subsection{Intensity Histograms}
\begin{figure}[!ht]
 \centering
 \includegraphics[width=0.85\textwidth,angle=0]{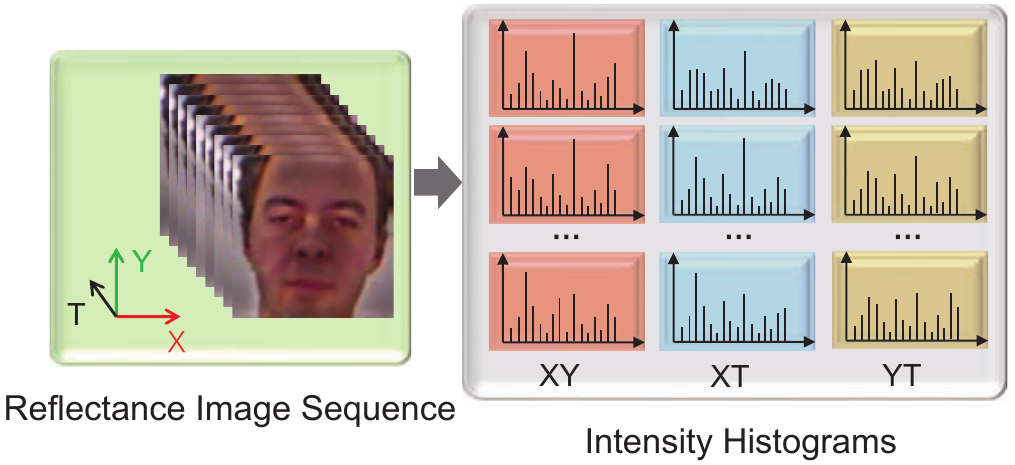}
 \caption{Extract intensity histograms from three orthogonal planes.}
 \label{fig:XYThistogram}
\end{figure}

In order to represent different intensities and distributions of reflectance images, we extract the intensity histograms from three orthogonal planes (i.e., XY, XT and YT) as shown in Fig. \ref{fig:XYThistogram}, where T represents the time dimension.

Considering the three orthogonal planes, the reflectance image sequence can be denoted as $\mathbb{R}=\{R_1,R_2,...,R_i,...R_n\}$, where $R_i$ denotes the $i$th reflectance image in $\mathbb{R}$ and $n$ is the length of the sequence. In our algorithm, we set $n=75$, which can capture intensity variations \cite{Boulkenafet2016FaceTIFS}. Before extracting intensity histograms, the size of color reflectance images is first empirically normalized to $150\times 150$. For the XY plane, intensity histogram from the color channels of each reflectance image of $\mathbb{R}$ is extracted and combined into a matrix, which is illustrated in Eq.\ref{Eq:HistogramMatrixXY}.
\begin{eqnarray}\label{Eq:HistogramMatrixXY}
  H_{XY}=
  \left [\begin{array}{c}
         h(R_{1}^{c1}), h(R_{1}^{c2}), h(R_{1}^{c3})\\
         h(R_{2}^{c1}), h(R_{2}^{c2}), h(R_{2}^{c3})\\
         ..., ..., ...\\
         h(R_{n}^{c1}), h(R_{n}^{c2}), h(R_{n}^{c3})
  \end{array}
  \right ]
\end{eqnarray}
where $c1$, $c2$ and $c3$ represent different color channels. $h(\cdot)$ means the operation of intensity histogram extraction and the dimension of $H_{XY}$ is $75\times 768$.

However, on XT and YT planes, we extract a slice of $\mathbb{R}$ every other row or column and count its intensity histograms also from different color channels. The reflectance image sequences from XT and YT can be written as $\mathbb{X}=\{X_1,X_2,...,X_i,...X_{\dot{n}}\}$ and $\mathbb{Y}=\{Y_1,Y_2,...,Y_i,...Y_{\ddot{n}}\}$, respectively. $\dot{n}$ and $\ddot{n}$ are the sequence length and equal to $n$. Finally, the XT and YT intensity histogram matrices with the size of $75\times 768$ are obtained based on Eq. \ref{Eq:HistogramMatrixXT} and Eq. \ref{Eq:HistogramMatrixYT}.

\begin{eqnarray}\label{Eq:HistogramMatrixXT}
  H_{XT}=
  \left [\begin{array}{c}
         h(X_{1}^{c1}), h(X_{1}^{c2}), h(X_{1}^{c3})\\
         h(X_{2}^{c1}), h(X_{2}^{c2}), h(X_{2}^{c3})\\
         ..., ..., ...\\
         h(X_{\dot{n}}^{c1}), h(X_{\dot{n}}^{c2}), h(X_{\dot{n}}^{c3})
  \end{array}
  \right ]
\end{eqnarray}

\begin{eqnarray}\label{Eq:HistogramMatrixYT}
  H_{YT}=
  \left [\begin{array}{c}
         h(Y_{1}^{c1}), h(Y_{1}^{c2}), h(Y_{1}^{c3})\\
         h(Y_{2}^{c1}), h(Y_{2}^{c2}), h(Y_{2}^{c3})\\
         ..., ..., ...\\
         h(Y_{\ddot{n}}^{c1}), h(Y_{\ddot{n}}^{c2}), h(Y_{\ddot{n}}^{c3})
  \end{array}
  \right ]
\end{eqnarray}

\begin{figure*}[!t]
\centering
\subfigure[Real face: XY histogram matrix]{\includegraphics[width=0.45\textwidth]{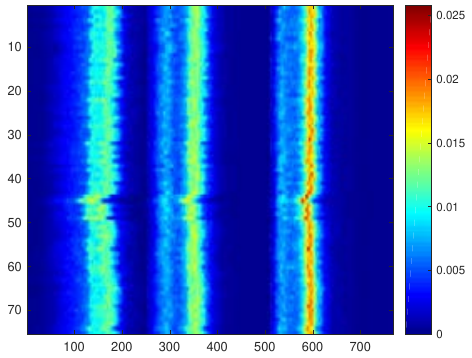}
\hfil
\label{fig:RealHist}}
\subfigure[Real face: Output of 1D CNN]{\includegraphics[width=0.45\textwidth]{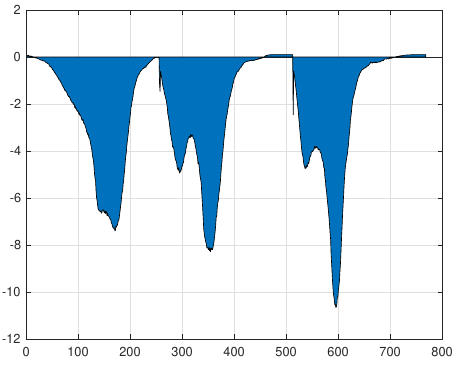}
\hfil
\label{fig:RealHistFea}}
\subfigure[3D mask: XY histogram matrix]{\includegraphics[width=0.45\textwidth]{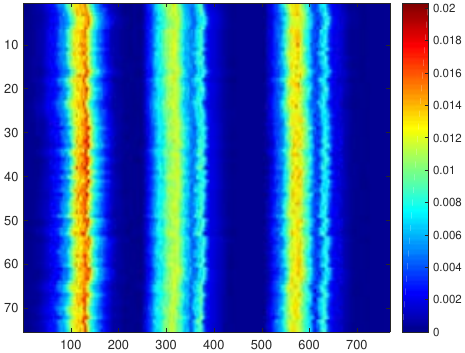}
\hfil
\label{fig:FakeHist}}
\subfigure[3D mask: Output of 1D CNN]{\includegraphics[width=0.45\textwidth]{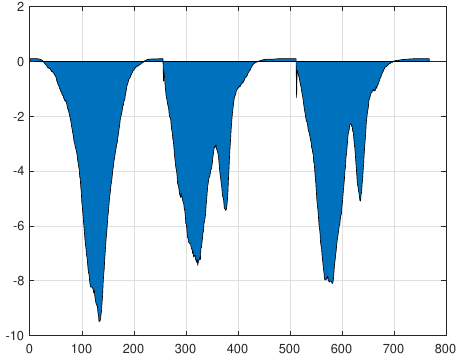}
\hfil
\label{fig:FakeHistFea}}
\caption{The input and output of 1D CNN network. (a) and (b) are the results of a real face reflectance image sequence, and (c) and (d) are the results of a 3D face mask reflectance image sequence. To make the histogram matrix better visualized, we use \textit{jet} colormap in (a) and (c). (b) and (d) plot the area of the output of 1D CNNs.}
\label{fig:HistFea}
\end{figure*}

\subsection{1D CNNs}
Different materials and surfaces react differently to illumination changes. Therefore, apart from intensity location features, we also use 1D CNN to extract intensity variation information from $H_{XY}$, $H_{XT}$ and $H_{YT}$. As shown in Fig. \ref{fig:FlowChart}, the 1D CNNs module presents three parallel 1D CNNs. In each network, 1D operation is employed to extract intensity variation information. Compared to other deep networks \cite{Yang2014Learn,Patel2016Cross}, the solution differs mainly in three aspects: (i) 1D convolutional filters are adopted to extract the temporal information of neighboring histograms; (ii) 1D pooling layers can effectively reduce the dimension after convolutional layers; (iii) The output features do not fuse the information between different intensity values.

The parameters of 1D CNN network are summarized in Table \ref{tab:Detail-CNN}. For all convolutional layers except the $16_{th}$ layer, the size of convolutional filters is set to $3\times1$. This means the convolutional filter can extract the intensity variation in three adjacent histograms and reduce network parameters as \cite{SajjadiSH2016EnhanceNet}. After each convolutional layer, we use Rectified Linear Units (ReLU) to activate the convolutional outputs \cite{Glorot2012Deep}. Then the pooling layers with size of $2\times1$ are utilized to gradually down-sample the results from ReLU layer. In the last pooling layer, the dimension of outputs is $1\times768$. Before feeding the outputs of last pooling layer into a fully connected (FC) layer, we introduce a transposition layer to reshape the size of outputs into $768\times 1$.

For face PAD, its essence is to classify whether the input is a real face or a presented 3D face mask. Thus, after FC layer, the most commonly used $SoftMax$ loss function is employed to train the network parameters \cite{Parkhi2015Deep}. In testing stage, we extract the outputs of the last Conv layer as the finally features of $H_{XY}$, $H_{XT}$ and $H_{YT}$ denoted as $C_{XY}$, $C_{XT}$ and $C_{YT}$. Fig. \ref{fig:HistFea} shows the intensity histogram matrixes and the learned intensity variation features of the reflectance image sequences of the real face and the 3D face mask in XY plane of RGB color space. After feature extraction, we concatenate $C_{XY}$, $C_{XT}$ and $C_{YT}$ into a feature vector with stronger representation, which are fed into an SVM classifier.

In training stage, the stochastic gradient descent (SGD) algorithm \cite{Bottou2010Large} is used to learn the network parameters. The learning rate is set to \(10^{-3}\). The momentum is set to 0.9 and weight decay 0.0005. In addition, we realize the 1D CNNs and SVM based on MatConvNet with the version 1.0-beta20 \footnote{http://www.vlfeat.org/matconvnet/} and liblinear with the version 1.96 \footnote{https://www.csie.ntu.edu.tw/$\sim$cjlin/liblinear/}, respectively.

\section{Experimental Setup}


\subsection{Experimental Data}
\begin{figure}[!ht]
 \centering
 \includegraphics[width=0.85\textwidth,angle=0]{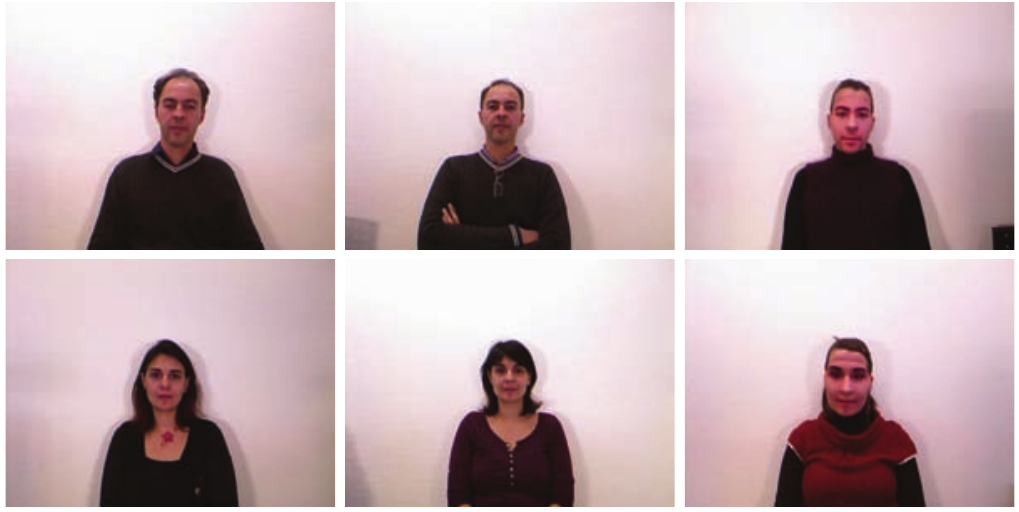}
 \caption{Samples from 3DMAD database. From left column to right column: real faces, real faces and 3D face mask.}
 \label{fig:3DMAD}
\end{figure}

\noindent\textbf{3D Mask Attack Database (3DMAD).} 3DMAD \footnote{https://www.idiap.ch/dataset/3dmad} \cite{Erdogmus2014Spoofing} is the first public-available database for 3D face mask PAD with 3D face masks from Thatsmyface\footnote{www.thatsmyface.com}. The database consists of 17 subjects, 3 sessions and total 255 video clips (76500 frames). Each subject has 15 videos with 10 real faces and 5 face masks. Videos are captured through Kinect sensors and are equipped with color and depth maps in $640 \times 480$ resolution. Fig. \ref{fig:3DMAD} shows some color examples of real face and 3D face mask.

\begin{figure}[!ht]
 \centering
 \includegraphics[width=0.85\textwidth,angle=0]{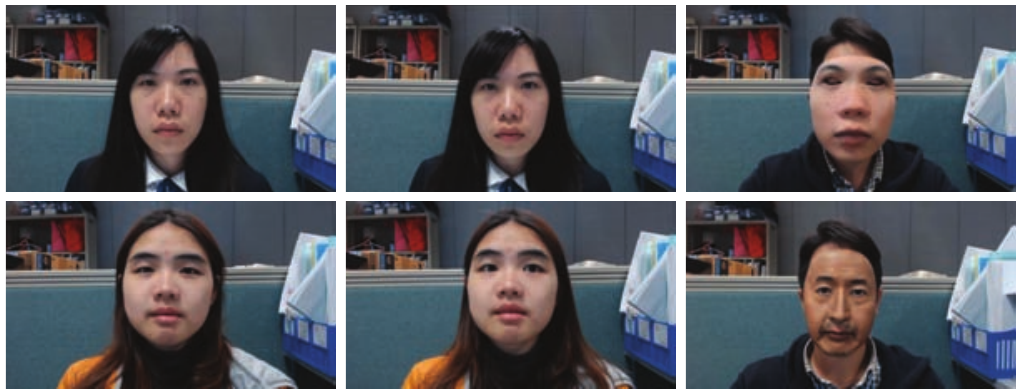}
 \caption{Samples from HKBU-MARs V1 database. From left column to right column: real faces, real faces and 3D face mask. For the mask, the first row represents the mask generated from Thatsmyface and the second row represents the mask generated from REAL-F.}
 \label{fig:HKBU}
\end{figure}

\noindent\textbf{HKBU-MARs V1.} HKBU-MARs V1 \footnote{http://rds.comp.hkbu.edu.hk/mars/} \cite{Liu20163D} is a supplementary database for 3D face mask PAD. It contains 120 videos from 8 subjects and masks. Two types of 3D masks are included: 6 from Thatsmyface and 2 from REAL-F \footnote{http://real-f.jp/en$\_$the-realface.html}. Each subject corresponds to 10 genuine samples and 5 masked samples. All videos are recorded through Logeitech C920 web-camera in $1280 \times 720$ resolution under room lighting condition. Fig. \ref{fig:HKBU} shows some color examples of real face and 3D face mask.

\begin{table*}[!ht]\scriptsize
\centering
\caption{The 3DMAD detection results of different plane combination mechanisms.}\label{tab:DifferentPlanes}
\begin{tabular}{|c|c|c|c|c|c|c|c|}
\hline
\multirow{2}{*}{Combined Planes} & \multicolumn{2}{c|}{Dev Set} & \multicolumn{5}{c|}{Test Set}              \\ \cline{2-8}
                        & EER(\%) & AUC(\%)   & APCER(\%) & BPCER(\%) & ACER(\%)     & EER(\%) & AUC(\%) \\ \hline \hline
XY                      & 2.2     & 99.74     & 2.8       & 5.0       & 3.9          & 0.0     & 100  \\ \hline
XT                      & 0.5     & 99.95     & 3.9       & 1.2       & 2.6          & 0.0     & 100  \\ \hline
YT                      & 0.4     & 99.98     & 2.4       & 2.5       & 2.4          & 0.0     & 100  \\ \hline
XY-XT                   & 0.8     & 99.94     & 2.7       & 2.8       & 2.7          & 0.0     & 100  \\ \hline
XY-YT                   & 0.1     & 100       & 1.2       & 3.2       & 2.2          & 0.0     & 100  \\ \hline
XT-YT                   & 0.0     & 100       & 0.1       & 0.8       & \textbf{0.4} & 0.0     & 100  \\ \hline
XY-XT-YT                & 0.0     & 100       & 0.2       & 1.2       & 0.7          & 0.0     & 100  \\ \hline
\end{tabular}
\end{table*}

\subsection{Evaluation Protocol}
For training and testing, we use the leave-one-out cross-validation (LOOCV) mechanism as used in paper \cite{Erdogmus2014Spoofing}. For 3DMAD database, one subject's data is left for testing while the data of the remaining 16 subjects is divided into two subject-disjoint halves as training and development sets in each fold of cross validations.
However, for KBU-MARs V1 database, we only use this database for inter-test as one person's video data is not published.

With performance evaluation, the results are reported in term of recently standardized ISO/IEC 30107-3 metrics \cite{ISO2016}: Attack Presentation Classification Error Rate (APCER) and Bona Fide Presentation Classification Error Rate (BPCER). It is noted that the APCER and BPCER depend on the decision threshold estimated by Equal Error Rate (EER) on the development set. To compare the overall system performance in a single value, the Average Classification Error Rate (ACER) is computed, which is the average of the APCER and the BPCER \cite{boulkenafet2017oulu}. To compare with existing works, we also report the results of EER and Area Under Curve (AUC) on the test set.

\section{Experimental Results and Discussion}
In this part, the intensity histograms of different planes are firstly evaluated on the detection performance. Then, various color spaces are analyzed for 3D face mask PAD. Furthermore, we extract the intensity histograms from original face images and compare the detection results with intrinsic images. After that, the inter-test experiments are performed on HKBU-MARs V1 database. Finally, the performance of our method is compared against the state-of-the-art approaches.

\begin{table*}[!ht]\scriptsize
\centering
\caption{The 3DMAD detection results of different color spaces.}\label{tab:DifferentColor}
\begin{tabular}{|c|c|c|c|c|c|c|c|}
\hline
\multirow{2}{*}{Color Space} & \multicolumn{2}{c|}{Dev Set} & \multicolumn{5}{c|}{Test Set}              \\ \cline{2-8}
                        & EER(\%) & AUC(\%) & APCER(\%) & BPCER(\%) & ACER(\%)     & EER(\%) & AUC(\%) \\ \hline \hline
RGB                     & 9.9     & 96.55   & 5.3       & 12.3      & 8.8          & 3.2     & 98.53  \\ \hline
YCbCr                   & 0.5     & 99.97   & 0.8       & 1.3       & 1.0          & 0.0     & 100    \\ \hline
HSV                     & 0.0     & 100     & 0.1       & 0.8       & \textbf{0.4} & 0.0     & 100    \\ \hline
\end{tabular}
\end{table*}

\subsection{Three Orthogonal Planes}
\begin{figure}[!ht]
 \centering
 \includegraphics[width=0.85\textwidth,angle=0]{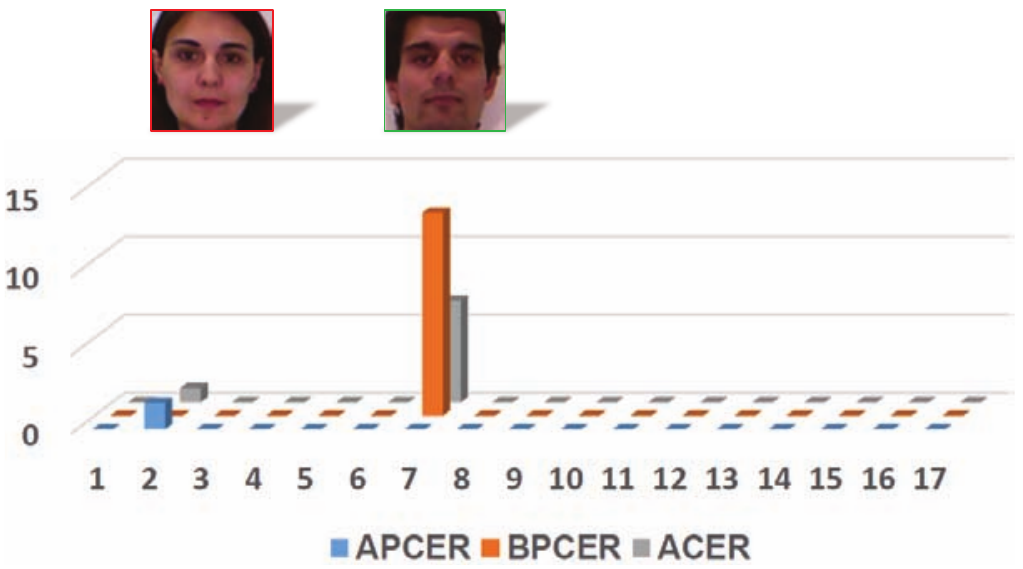}
 \caption{The APCER, BPCER and ACER of each subject in 3DMAD. The horizontal axis represents the subject's ID.}
 \label{fig:Excel}
\end{figure}

Table \ref{tab:DifferentPlanes} illustrates the results of the intensity histograms extracted from different planes and combined using different mechanisms. From the table, we can clearly find that all averaged EERs and all averaged AUCs of test sets are 0.0\% and 100\%. This means that our proposed method can accurately distinguish between real faces and 3D masks by the decision threshold determined by the data set itself. However, when the test set uses the threshold of the development set, the detection performance of the algorithm in the test set will decrease. In addition, for combining different intensity histograms, it is shown that simultaneously concatenating TOP histograms will not improve the detection accuracy. We conjecture that the reason may lie in that the feature dimension is too high for the classifier to overfit. Rather, when cascading the intensity histograms extracted from XT and YT planes, our algorithm achieve the best results with APCER=0.1\%, BPCER=0.8\% and ACER=0.4\%. This reflects the fact that 3D face masks are more sensitive to the changes in lighting than real face skin. Fig. \ref{fig:Excel} shows the detailed classification of the best case in LOOCV mechanism. For instance, there are some classification failures in the samples of the $2$nd and $7$th subjects.

\begin{table*}[!ht]\scriptsize
\centering
\caption{Compare our proposed method with original face images.}\label{tab:CompareOriginalFace}
\begin{tabular}{|c|c|c|c|c|c|c|c|}
\hline
\multirow{2}{*}{Face Image} & \multicolumn{2}{c|}{Dev Set} & \multicolumn{5}{c|}{Test Set}              \\ \cline{2-8}
                        & EER(\%) & AUC(\%) & APCER(\%) & BPCER(\%) & ACER(\%)     & EER(\%) & AUC(\%) \\ \hline \hline
Original Face           & 0.7     & 99.98   & 1.6       & 0.6       & 1.1          & 0.0     & 100     \\ \hline
Intrinsic Face          & 0.0     & 100     & 0.1       & 0.8       & 0.4          & 0.0     & 100     \\ \hline
\end{tabular}
\end{table*}

\begin{table*}[!ht]\footnotesize
\centering
\caption{The detection results of inter-test.}\label{tab:CrossTest}
\begin{tabular}{|c|c|c|c|c|c|}
\hline
\multirow{2}{*}{Face Mask} & \multicolumn{5}{c|}{Trained on 3DMAD Test on HKBU}   \\ \cline{2-6}
                           & APCER(\%) & BPCER(\%) & ACER(\%) & EER(\%) & AUC(\%)  \\ \hline \hline
Thatsmyface                & 100       & 0         & 50       & 57.2    & 49.99    \\ \hline
REAL-F                     & 100       & 0         & 50       & 89.8    & 4.4      \\ \hline
\end{tabular}
\end{table*}

\subsection{Different Color Spaces}
Furthermore, we extract the histograms from different color spaces (i.e. RGB, YCbCr and HSV), and compare the detection results in Table \ref{tab:DifferentColor}. The table shows that the color is a very important clue for 3D face mask PAD. Compared to the RGB color space, the YCbCr and HSV display color and brightness in different channels. For instance, the HSV shows the brightness information in H channel and the color information in S and V channels. When the intensity histograms are extracted from the color channel and brightness channel respectively, the detection performance of our algorithm can be significantly improved. It can be observed in Table \ref{tab:DifferentColor} that the ACER of HSV is 22 times lower than the ACER of the RGB color space.


\subsection{Comparison with Original Face Image}
In order to verify the validity of the proposed intrinsic image analysis, we extract the intensity histograms from original face images in HSV color space and calculate the intensity variational information also using 1D CNN. The comparison results are shown in Table \mbox{\ref{tab:CompareOriginalFace}}. When extracting the intensity histograms from original face images, the EER of dev set and ACER of test set are 0.7\% and 1.1\%, respectively. Compared the results of our proposed intrinsic image analysis based method, the ACER is reduced by nearly three times. This shows that it is reasonable and effective to detect 3D face mask in intrinsic image space.

\subsection{Cross Database Evaluation}
To gain insight into the generalization capabilities of our proposed detection method, we conduct a cross-database evaluation. To be more specific, the countermeasure is trained and tuned on 3DMAD and then tested on HKBU-MARs V1 database.
The obtained detection results are summarized in Table \ref{tab:CrossTest}. During the experiment, the first 9 subjects in 3DMAD are used to train our algorithm, and the remaining 8 subjects are used as the development set. When classifying according to the threshold of 3DMAD database, all 3D face masks are detected incorrectly with APCER=100\% and all real faces are correctly classified with BPCER=0\%. The reason may lie in that the materials of face mask of these two databases are different. From these results, we conclude that our proposed method is only for detecting the 3D face masks with specific material.


\begin{table}[!ht]\footnotesize
\centering
\caption{Comparison between our proposed countermeasure and state-of-the-art methods on 3DMAD database.}\label{tab:Compare}
\begin{tabular}{|c|c|c|c|}
\hline
\multirow{2}{*}{Methods} & \multicolumn{3}{c|}{Test Set}              \\ \cline{2-4}
                                         & ACER(\%)     & EER(\%)      & AUC(\%) \\ \hline \hline
MS-LBP \cite{Erdogmus2014Spoofing}$\dag$ & -            & 5.2          & 98.65 \\ \hline
fc-CNN \cite{Sun2007Blinking}$\dag$      & -            & 3.2          & 98.36 \\ \hline
LBP-TOP \cite{Pereira2014Face}$\dag$     & -            & 1.4          & 99.92 \\ \hline
rPPG\cite{Liu20163D}$\dag$               & -            & 8.6          & 96.81 \\ \hline
DCCTL \cite{Shao2017Deep}                & -            & 0.6          & 99.99 \\ \hline
Pulse \cite{Li2017Generalized}           & 7.9          & 4.7          & - \\ \hline \hline
Our Method                               & \textbf{0.4} & \textbf{0.0} & \textbf{100} \\ \hline
\end{tabular}
\begin{tablenotes}
    \item $\dag$ the results are reported in \cite{Shao2017Deep}.
\end{tablenotes}
\end{table}


\subsection{Comparison with the State of the Art}
Table \ref{tab:Compare} gives a comparison with the state-of-the-art in 3D face mask PAD techniques proposed in the literature. It can be seen that our proposed intrinsic image analysis based method outperforms the-state-of-the-art algorithms on the EERs of development set test set. Especially compared with the pulse detection based algorithms \cite{Liu20163D, Li2017Generalized}, the ACER of our proposed method is 0.4\%, which is nearly 20 times better than the ACER obtained by \cite{Li2017Generalized}. As aforementioned, the reason may lie in the pulse detection based methods are sensitive to camera settings and light conditions. For LBP-TOP \cite{Pereira2014Face} and DCCTL \cite{Shao2017Deep}, their EERs are 1.4\% and 0.6\%, respectively. Even though the performances are better than pulse detection based methods, they can be interfered by the attacker's head shake. Unfortunately, for 3D face mask PAD, there is only one fully public database (i.e. 3DMAD) so far, which makes it difficult to fully verify the effectiveness of our algorithm.

\section{Conclusion}
In this article, we addressed the problem of 3D face mask PAD from the viewpoint of intrinsic image analysis. We designed a new feature for reflectance characteristic description. Apart from that, we also introduced 1D CNN to extract intensity variation features. Extensive experiments on 3DMAD database showed excellent results. However, deep features extracted from three orthogonal planes should have different weights, but in our approach we simply concatenate them together. Therefore, we will design a attention model to tackle this problem. For the generalization capability, we think developing a 3D face mask PAD method with interoperability is big open issue, which is far from the current state of the art. So we leave it to future work.

\section*{References}

\bibliography{spoofingNew2}
\end{document}